\documentclass[runningheads]{llncs}

\usepackage{eccv}

\usepackage{eccvabbrv}

\usepackage{multirow}
\usepackage{multicol}
\usepackage{graphicx}
\usepackage{booktabs}
\usepackage{algorithm}
\usepackage{algpseudocode}
\usepackage{enumitem}

\usepackage[accsupp]{axessibility}  
\def\xx{\mathbf{x}}
\def\ee{\mathcal{E}}
\def\cc{\mathcal{C}}
\def\pp{\mathcal{P}}

\def\epsbia{\varepsilon_{\text{bia}}}
\def\kk{K_{\text{bia}}}
\newcommand{\parallelsum}{\mathbin{\|}}

\usepackage{orcidlink}

\begin{document}

\title{Debiasing surgeon: fantastic weights and how to find them} 

\author{Rémi Nahon\orcidlink{0009-0009-0757-5716} \and
Ivan Luiz De Moura Matos\orcidlink{0009-0009-5032-0444}\and
Van-Tam~Nguyen~\and~Enzo~Tartaglione\orcidlink{0000−0003−4274−8298}}

\authorrunning{R.~Nahon et al.}

\institute{LTCI, Télécom Paris, Institut Polytechnique de Paris, France 
\email{\{name.surname\}@telecom-paris.fr}}

\maketitle

\begin{abstract}
Nowadays an ever-growing concerning phenomenon, the emergence of algorithmic biases that can lead to unfair models, emerges. Several debiasing approaches have been proposed in the realm of deep learning, employing more or less sophisticated approaches to discourage these models from massively employing these biases. However, a question emerges: is this extra complexity really necessary? Is a vanilla-trained model already embodying some ``unbiased sub-networks'' that can be used in isolation and propose a solution without relying on the algorithmic biases? In this work~\footnote{Published in the European Conference on Computer Vision 2024}, we show that such a sub-network typically exists, and can be extracted from a vanilla-trained model without requiring additional fine-tuning of the pruned network. We further validate that such specific architecture is incapable of learning a specific bias, suggesting that there are possible architectural countermeasures to the problem of biases in deep neural networks. 
\end{abstract}

\section{Introduction}
\label{sec:introduction}

In the last decade, recent technical and technological advances enabled the large-scale deployability of deep learning (DL)-based approaches, impacting, among others, the computer vision community. The possibility of training systems in an end-to-end fashion enables access to non-trivial solutions to complex tasks, ushering in unprecedented breakthroughs and fundamentally reshaping the landscape of visual perception. The transformative impact of deep learning finds nowadays broad applicability in diverse real-world scenarios, including autonomous driving, medical imaging~\cite{ma20233d}, augmented reality~\cite{li2020object}, and robotics~\cite{cebollada2021state}. As the computer vision community continues to harness the power of deep learning with scaling models and methods, combining for example language and vision models~\cite{xu2023multimodal, fan2024improving}, the boundaries of what is achievable in visual understanding are continually pushed, promising exciting avenues for innovation and discovery.
\begin{figure}
    \centering
    \includegraphics[width=0.7\linewidth]{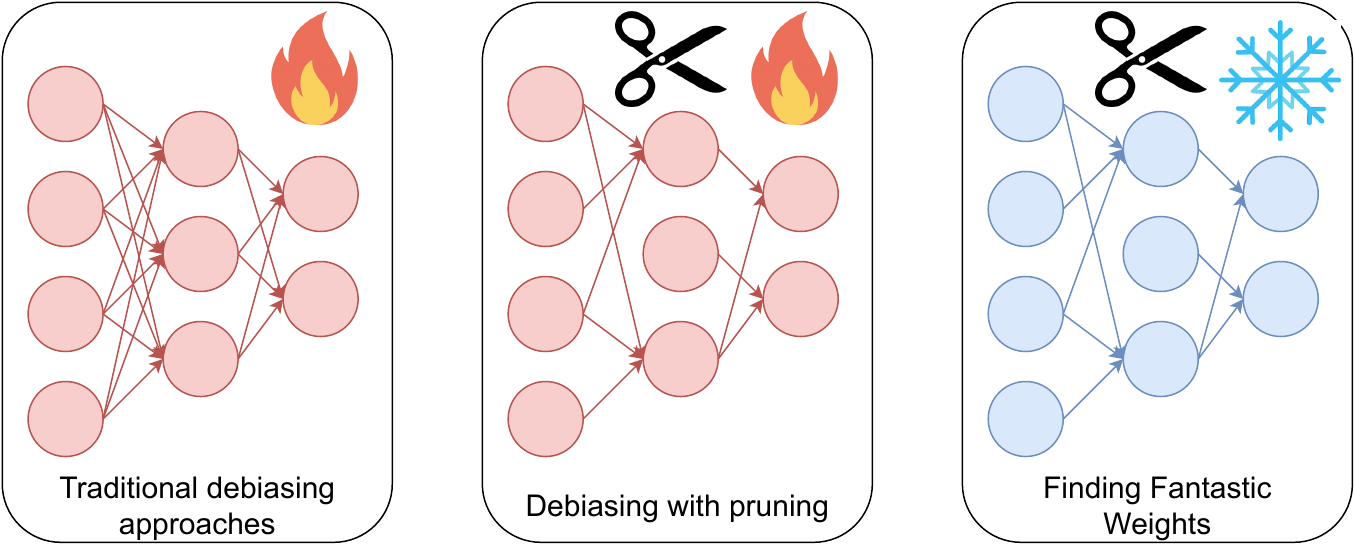}
    \caption{Despite other debiasing approaches implying training or fine-tuning the whole model, with Finding Fantastic Weights (FFW) we maintain the model's parameters frozen and remove the sub-network responsible for bias information propagation.}
    \label{fig:teaser}
\end{figure}

Unfortunately, from big power comes big responsibility: a big aspect to account for comes from the need to avoid the unintended over-reliance on spurious correlations or biases, naturally present in datasets~\cite{izmailov2022feature}. This poses a practical significant challenge in DL real-world deployment. As an explicative case, in the context of image classification tasks such as detecting pedestrians, if environmental cues (e.g., the presence of a sidewalk/pedestrian crossing) become spuriously correlated with the target classes, neural networks may exploit such correlations as \emph{shortcuts} for classification~\cite{geirhos2020shortcut}, thereby leading to performance degradation when presented with images containing different backgrounds (e.g., pedestrian crossing the road not on pedestrian crossing lanes). This leads to some potential threats in certain applications.

In 2021, the European Commission put forward the Artificial Intelligence Act (AI Act)~\cite{aiact}, aiming to categorize and oversee artificial intelligence implementations according to their potential to cause harm~\cite{veale2021demystifying, hupont2023documenting}. Similarly to the General Data Protection Regulation  (GDPR)~\cite{regulation2018general}, the AI Act has the potential to establish a global benchmark~\cite{aiact}. Regardless, the EU's AI regulation is already garnering attention worldwide: for example, in 2021 Brazil's Congress approved a bill to establish a legal framework for artificial intelligence, pending approval by the country's Senate. Guaranteeing the avoidance of spurious biases that might undermine the safety, trust, and accountability of DL models is then not only becoming a matter of safety but will soon be a legal constraint.

Multiple solutions are proposed in the last luster, all relying on relatively heavy re-training approaches. More specifically, in the realm of DL debiasing, we can identify three main lines of research: (i) supervised, where the labels of the bias are provided; (ii) bias-tailored, where proxy models can capture specific biases; (iii) unsupervised, where biases are guessed directly within the vanilla model. All three approaches typically require heavy training procedures for either properly tuning the hyper-parameters of the models, or training multiple times the neural networks. This comes at another very relevant environmental cost, and besides it is not granted that such solutions can always apply to any architecture/task. Some recent studies hinder the possibility of reusing pre-trained models and removing bias sources with some adjustments~\cite{zhang2022correct, zhao2021learning}, but none of them provides a final solution yet.

In this work we show the existence of unbiased sub-networks in vanilla-trained models, providing also some analysis on the final performance. In a nutshell, we learn how to mask trained weights such that the information about the bias becomes non-extractable by the layer(s) entitled to solve the target task. Therefore, we {\bf F}ind the {\bf F}antastic {\bf W}eights (FFW) constituting an unbiased sub-network that solves the tasks, without further fine-tuning (Fig.~\ref{fig:teaser}). Our results, besides providing a clear indication that such a sub-network exists in vanilla-trained overfitting models, contribute to the DL debiasing community, showing that these solutions are achievable without relying on over-complex training schemes.

At a glance, this work proposes the following contributions:
\begin{itemize}[noitemsep, nolistsep]
    \item We draft a theory that shows how the performance on some given task might depend on biased features, suggesting that removing the bias source \emph{not always} results in an enhanced performance (Sec.~\ref{sec:theory});
    \item To our knowledge, FFW is the first parameter selection strategy, from a vanilla-trained model and without the model's retraining, that provides some guarantees on the bias-related extractable information from the task classifier (Sec.~\ref{sec:ffw});
    \item We show that the non-spurious correlations are learned even in vanilla-trained models, in later training stages (Sec.~\ref{sec:fitting}), suggesting that over-complicated learning setups are not needed for debiasing;
    \item We test on very common setups for the community, showing the existence of these sub-networks (Sec.~\ref{sec:mainresults}). This opens the road to the development of the design of more energy-efficient debiasing strategies.
\end{itemize}
\section{Related works}
\label{sec:related}

\textbf{Spurious correlations.} The risk of the insurgence of spurious correlations employed for a given target task is a broadly acknowledged issue in the DL community. This phenomenon is known as \emph{shortcut learning}~\cite{geirhos2020shortcut}, where the algorithm relies on the simplest correlations to fit the training set, but do not generalize well. Some of these can range from the texture of images~\cite{geirhos2018imagenet} to biases inherent in language~\cite{gururangan2018annotation}, or even sensitive variables like ethnicity or gender~\cite{narayanan2018translation, feldman2015certifying}. Such behavior raises practical concerns, particularly in applications where the reliability of DL is critical, such as healthcare, finance, and legal services~\cite{fairness_metrics}.\\
\textbf{Debiasing.} Recent efforts aimed at training debiased networks resilient to spurious correlations can be broadly classified into two major categories. \\
Some approaches rely on some information related to the nature of the presence of biases, and we will name these as \emph{supervised} approaches. Among these, we mention \emph{data sanitizing} approaches, working as \emph{pre-processing} approaches, employing for example GANs~\cite{choi2020stargan, kang2021content} or style-transfer~\cite{geirhos2018imagenet}, \emph{post-processing} methods attempting to correct biased behaviors~\cite{hardt2016equality, kamiran2012decision} and \emph{in-processing} methods, trying to sanitize the model directly at training time~\cite{rebias, end, nam2020learning}. \\
Besides these, a new class of \emph{unsupervised} approaches is rising~\cite{nahon}, where little or even no clue on the nature of the bias is provided. Specifically, we can recognize models pre-capturing texture biases~\cite{wang2018learning, goel2020model}, leveraging groups imbalances~\cite{Debian, EIIL, arjovsky2019invariant} and assuming that the bias is learned in early training stages~\cite{nam2020learning, lee2021learning}.\\
\textbf{Studying impacts of neural architectures.} Recently a lot of attention has been devoted in studying the influence of a deep neural network's architecture on its generalizability. For example, Diffenderfer~\etal~\cite{diffenderfer2021winning} built a pruning algorithm inspired by the lottery ticket hypothesis~\cite{frankle2018lottery} to craft compact and resilient network architectures. Similarly, Bai~\etal~\cite{bai2021ood} tackles a similar issue through the lenses of out-of-distribution performance. This class of approaches, although effective at the cost of enhanced computational complexity~\cite{tartaglione2022rise}, does not entirely eradicate connections to spurious attributes. On the other hand, Zhang~\etal and Zhao~\etal suggest the efficacy of pruning weights associated with such attributes sided with some fine-tuning~\cite{zhang2021can, rest}, finding some counter-arguments in the literature~\cite{blakeney2021simon} and even with works suggesting the need to employ corrective distillation terms~\cite{o2022self}. The discussion of these empirical findings is still open.
\\In a recent work, it has been empirically showcased the presence of sub-networks within neural networks that exhibit reduced susceptibility to spurious features~\cite{zhang2021can}. By capitalizing on the modular nature inherent in neural architectures~\cite{csordas2020neural}, Zhang~\etal were able to demonstrate that, in principle, it is possible to select, already at initialization, networks that will avoid the employment of biased features. However, this is not a sufficient condition to still guarantee the presence of such sub-networks in an already-trained model. In the era of foundation models, indeed, training a model from scratch can be very costly and impractical~\cite{zayed2023fairness}. In this work we focus on this problem, characterizing the condition of extractability for such debiased sub-networks, drawing as well a link between bias removal and impact on the performance of the target task. We will sketch the theory motivating our approach, and we will study the problem of extracting these sub-networks, trained with vanilla strategies, \emph{without requiring finetuning}.
\section{Method}
\label{sec:method}

\subsection{Removing the bias impacts the performance}
\label{sec:theory}
In this section, we will define the problem of learning a mapping between the input $\xx$ and its target output $\hat{y}$ on a given dataset $\mathcal{D}$ (that is traditionally slit in train, validation, and test), under the assumption that there is an underlying bias $\hat{b}$ (namely, ``spurious'' correlations) associated to it. We can model this problem working with the random variables associated with these quantities: we define $\hat{Y}$ as the random variable associated with the target output, $Y$ as the random variable associated with the output of the DL model composed of an encoder $\ee$ and a classifier $\cc$, $\hat{B}$ as the random variable associated to the bias, and $B$ the random variable associated to the extractable information regarding the bias, from $\ee$. We can easily write the joint probability between $\hat{Y}$ and $Y$:
\begin{equation}
    p\left(Y,\hat{Y}\right) = \frac{\scriptstyle 1}{\scriptstyle N_C}\left[\delta_{y\hat{y}}(1-\varepsilon_{y\hat{y}}) + \bar{\delta}_{y\hat{y}}\frac{\scriptstyle \varepsilon_{y\hat{y}}}{\scriptstyle N_C-1}\right]
\end{equation}
where $\varepsilon_{y\hat{y}}$ indicates the classification error on $N_C$ classes, $\delta_{ab}$ is the Kronecker delta, and $\bar{\delta}_{ab}=1-\delta_{ab}$. In debiasing problems, we can also write the joint probability between $\hat{Y}$ and $\hat{B}$ as
\begin{equation}
    \label{eq:basicbias}
    p\left(\hat{Y}, \hat{B}\right) = \frac{\scriptstyle 1}{\scriptstyle N_C}\left[\delta_{\hat{y}\hat{b}} \rho + \bar{\delta}_{\hat{y}\hat{b}}\frac{\scriptstyle 1-\rho}{\scriptstyle N_B-1}\right]
\end{equation}
where $N_B$ is the number of biases and $\rho$ is the correlation between bias and target. For the sake of tractability, the bias is in \eqref{eq:basicbias} presented to be uniformly distributed across the classes, but in general this is not true (it is sufficient to add an index to the specific bias-target class pairing for $\rho$). In this case, if $~{\rho=\frac{1}{N_B}}$, we know the dataset $\mathcal{D}$ is balanced wrt. the bias; otherwise, this could lead to the employment, from $\cc$, of spurious features, which can imply generalization and fairness issues. Indeed, we can write the mutual information between the target class and the bias as
\begin{equation}
    \mathcal{I}(\hat{B}, \hat{Y}) =   \frac{\scriptstyle N_B}{\scriptstyle N_C}\left[\log(N_B) +\rho \log(\rho) + (1-\rho)\log\left(\frac{\scriptstyle 1-\rho}{\scriptstyle N_B-1}\right)\right].
\end{equation}
The full derivation can be found in the Appendix. Evidently, there is an implicit link between bias and target classes. Considering that we aim at learning a mapping between input and output, we can introduce the implicit parameter $\phi$ that encodes the tendency of our model to rely on biased features: we call this parameter \emph{model's biasedness} and it contributes as well to the error $\varepsilon$. For tractability, we will now treat the problem assuming one bias source per target class (the same is extendible to having more bias sources per target class, as well as having one same bias per multiple target classes), and for simplicity, we have $~{N_C=N_B=N}$. Besides, we will assume that all the sources of error are due to the presence of the bias and that they are uniform across classes. It is possible to write a joint probability distribution between $\hat{B}, \hat{Y}, Y$, and by marginalizing over $\hat{Y}$, we can obtain
\begin{equation}
   \resizebox{.99\hsize}{!}{ $\mathcal{I}(\hat{B}, Y) = \frac{\scriptstyle \rho + (1-\phi)(1-\rho)}{\scriptstyle \log(N)}\log\left[\rho + \phi(1-\rho)\right]+ \frac{\scriptstyle \phi(1-\rho)}{\scriptstyle \log(N)} \log \left [\frac{\scriptstyle (1-\phi)(1-\rho)}{\scriptstyle N-1} \right ]+ 1 .$}
\end{equation}

The full derivation can be found in the Appendix. During debiasing, what we would like to achieve, is to minimize such a quantity, which is achieved for $\phi\rightarrow 0$: on the contrary, we can evidently see that the model is completely biased when $\phi\rightarrow 1$. \\
Changing the biasedness also affects the predicted performance, and it is possible that by completely removing the information regarding the bias, the overall performance of the classifier will be harmed. Specifically, let us define the error disentangled from the bias as $\varepsilon_{\text{unb}}=0$ and the error due to the lack of the bias features as
\begin{equation}
    \varepsilon_{\text{bia}} = \kk(1-\phi),
\end{equation}

where $\kk$ is the inherent ground-truth dependence between the use of the biased feature and the target label, and is an implicit property of the learned task that we name \emph{task's biasedness}. Specifically, $\kk$ indicates how much the target task depends on the bias to be solved. We can comprehensively write the mutual information between the target label and the model's prediction as
\begin{align}
    \mathcal{I}&(\hat{Y}, Y) = 
    f\left\{
    {\displaystyle \frac{\scriptstyle 1}{\scriptstyle N}\left[\rho(1- \epsbia) + (1-\phi)(1-\rho)(1-\kk)\right], N,N^2}
    \right\} 
    + \nonumber\\
    &f\left\{
    {\displaystyle \frac{\scriptstyle 1}{\scriptstyle N}\left[\frac{\scriptstyle \phi(1-\rho)}{\scriptstyle N-1}\kk + \left(\frac{\scriptstyle \rho^2(N-2) + \rho(1-\rho)(N-2)}{\scriptstyle (N-2+\rho)}\right)\epsbia\right], N(N-1), N^2}
    \right\},\label{eq:MIdependence}
\end{align}
where $f(x,y,z) = x y \log(x z)$.
\begin{figure}[t]
    \centering
    \begin{minipage}[b]{0.3\textwidth}
        \centering
        \begin{subfigure}{\linewidth}
            \centering
            \includegraphics[width=\linewidth]{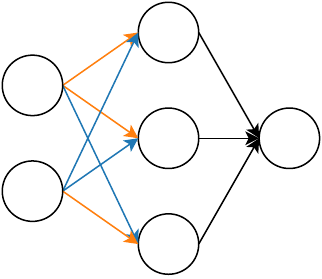}
            \caption*{\bf (a)}
            \label{fig:first}
        \end{subfigure}
        
        \begin{subfigure}{\linewidth}
            \centering
            \includegraphics[width=\linewidth]{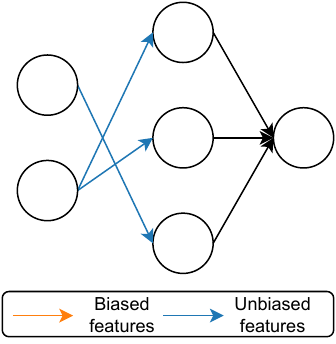}
            \caption*{\bf (b)}
            \label{fig:second}
        \end{subfigure}
    \end{minipage}%
    \begin{minipage}[b]{0.6\textwidth}
        \centering
        \begin{subfigure}{\linewidth}
        \centering
        \includegraphics[width=\linewidth]{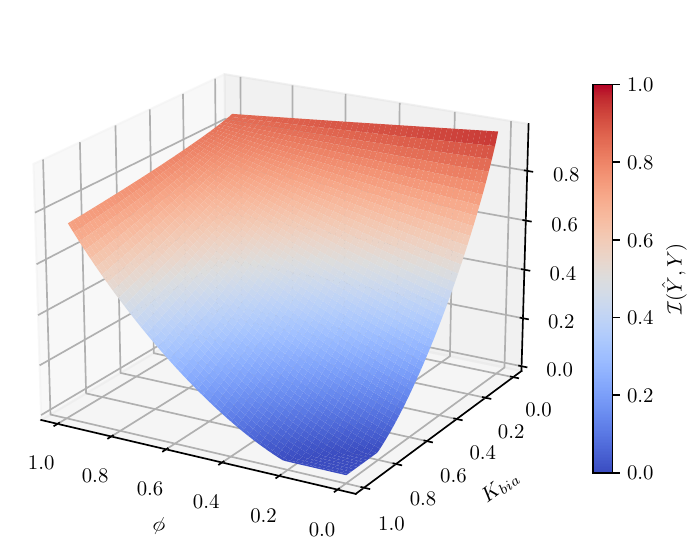}
        \caption*{\bf (c)}
        \label{fig:third}
        \end{subfigure}
    \end{minipage}
    \caption{The vanilla model, where the model still employs information related to the bias ($\phi\neq 0$)~(a), the model where the information of the bias is entirely removed ($\phi=0$)~(b), and the relationship between model biasedness $\phi$ and task biasedness $\kk$ (plot obtained with $N=10$ and $\rho=0.9$) (c).}
    \label{fig:set_alpha}
\end{figure}

The full derivation for \eqref{eq:MIdependence} can be found in the Appendix. Fig.~\ref{fig:set_alpha}c displays how the performance of a predictor drops as a function of both $\kk$ and $\phi$. As we can observe, there is a clear dependence between the remotion of the bias features ($\phi\rightarrow 0$) and the final model's performance, depending on the dataset-related parameter $\kk$; if $\kk$ is high, then the performance drops as $\phi$ approaches zero; on the contrary, if $\kk$ is low, the performance increases as $\phi\rightarrow 0$. This tells us that debiasing, intended as fair use of features without biases, does not imply necessarily \emph{removal} of these, but rather \emph{finding a proper balancing} between the employed features. To achieve this, one possibility is, for example, to mask the information related to the bias, at the output of $\ee$. Intuitively, if the information related to the bias can be disentangled from the target, this can be effectively achieved - however, if the target task intrinsically requires the employment of the bias information, then its effective removal will harm the target task performance. In the next section, we will try to unveil what is the effect of removing the information related to the bias in a structured way.

\subsection{Towards bias information removal}

Following on the previous finding, we here analyze a na\"ive approach that allows in principle to solve the problem of removing biases in a DL model. From~\cite{poole2019variational}, we know that we can upper bound the mutual information between the extractable information about the bias at the output of $\ee$ according to
\begin{equation}
    \label{eq:upper}
    \mathcal{I}(\hat{B},B) \leq \log \frac{\scriptstyle \exp(s^{\parallelsum})}{\frac{1}{N_B-1}{\scriptstyle \sum_j \exp(s_j^\perp)}},
\end{equation}
where $s$ is a similarity measure, typically expressed as \emph{cosine similarity}, in the feature space (in our case, the output of $\ee$), between samples having same \emph{bias aligned} ($s^{\parallelsum}$), and \emph{bias misaligned} samples ($s^\perp$). What we aim to achieve through debiasing approaches is in general to minimize \eqref{eq:upper}: we can achieve it by either minimizing the numerator (maintaining the denominator constant) or maximizing the denominator (maintaining the numerator constant). Some works in the literature attempted to tackle this problem directly, designing proper regularization or loss functions, requiring training of the model~\cite{end, hong2021unbiased, barbano2023unbiased}.

One intriguing possibility, as already suggested in a recent work~\cite{zayed2023fairness}, is to employ \emph{structured pruning} to completely mask all the layer's outputs that are responsible for biased information flow. More formally, let us name \emph{bottleneck layer} the output of the encoder $\ee$, producing a vector $\mathbf{z}\in \mathbb{R}^{N_{\text{bott}}}$, where $N_{\text{bott}}$ is the output size of the bottleneck layer. Assuming that the information regarding the bias is encoded by a subset of these neurons, it is in principle possible to find a pruning mask $\mathcal{M}$ such that the output $\hat{\mathbf{z}}= \ee(\mathbf{x}, \mathbf{w}) \odot \mathcal{M}$ does not encode the information related to the bias. This approach is in principle not always destined to succeed, and on the contrary, without an explicitly constrained training procedure that enforces the disentanglement between bias and target features, it is unlikely to be successful (as in Fig.~\ref{fig:set_alpha}a, it is not possible to mask neurons holding information of the bias and at the same time solve the task). It is indeed possible that the information about the bias is distributed across multiple dimensions of $\mathbf{z}$. Furthermore, given the non-linearity relationship inherently encoded in $\ee$ between input and its output, any linear mapping, including approaches like the selection of Principal Components, is potentially unsuitable for the purpose. One more viable and flexible approach would be to perform \emph{unstructured pruning}, sanitizing the DL model from the propagation and the computation of any information regarding the bias (as also visualized in Fig.~\ref{fig:set_alpha}b). There are indeed some weights encoding the information related to the bias, and there are \emph{fantastic weights} that do not process such information. In the next section, we will discuss how to find them.

\subsection{Finding the Fantastic Weights}
\label{sec:ffw}
In this section, we will provide the basics regarding how to find a mask $\mathcal{M}$ on atrained DL model that removes the information related to biases. Overall, we will assume that a vanilla-trained model, composed of an encoder $\mathcal{E}$ and a task classifier head $\mathcal{C}$, is provided. We will not require any constraint regarding the way such a model is trained, and the parameters belonging to those two blocks will not be modified during the process (apart from allowing their pruning). \\
\textbf{Estimating bias leaking information.} To estimate the amount of bias information leaking to the classifier $\mathcal{C}$, we will attach at the output of the encoder (the \emph{bottleneck layer}) an auxiliary bias extraction head $\mathcal{P}$ coined Privacy Head in~\cite{irene} its purpose was ``[to estimate] the information we desire to maintain private'': the bias . We will design $\mathcal{P}$ such that it has the same number of parameters as $\mathcal{C}$: this head will be trained to estimate the possible information that can be used by the classifier about the bias. To this end, we can train $\mathcal{P}$ using a classification loss, like the categorical cross-entropy loss (CCE). Along the process, we will not allow the error signal to backpropagate through the encoder $\mathcal{E}$. \\
\textbf{Metrics to remove the bias.} We want to minimize the achievable performance of $\mathcal{P}$. Some works suggest that techniques like gradient inversion~\cite{LNL} are potentially employable in such a context; however, in this case, such an approach is not a good fit. More specifically, maximizing CCE does not really remove information, but simply minimizes the activation of the correct class (as also explained in some works like~\cite{nahon}). What we can on the contrary employ, is to minimize the mutual information between bias labels and the output of $\mathcal{P}$, making the output converge as much as possible to a uniform distribution. Such a term will not contribute to the update of $\mathcal{P}$ but it will reach the parameters in $\mathcal{E}$, to be then used to estimate the pruning masks.\\
\textbf{Estimating the bias mask.} In FFW we propose two possible approaches to find the pruning masks: one acts unstructuredly, while in the other we will behave in a structured manner. In an unstructured sub-network selection setup, every parameter in $\mathcal{E}$ will have a gating parameter $m_i\in \mathbb{R}$ associated with it, initialized to zero. The value of each parameter $w_i$ will be replaced by
\begin{equation}
    \label{eq:hatw}
    \hat{w}_i = w_i\cdot \left[\Theta(-m_i)\cdot 2\sigma\left(\frac{m_i}{\tau}\right) + \Theta(m_i)\right],
\end{equation}
where $\tau$ is a temperature, $\sigma$ is the sigmoid function, and $\Theta(x)$ is the one-step function. To maintain the differentiability of our formulation, at the differentiation stage, we employ a straight-through estimator~\cite{bengio2013estimating} for $\Theta$. As $\tau\rightarrow 0^+$, the value of \eqref{eq:hatw} will become either $w_i$ (unpruned) or $0$ (pruned). In the structured variant, the whole unit will be pruned, and the $j$-th neuron's output $z_j$ is masked, according to
\begin{equation}
    \label{eq:hatz}
    \hat{z}_j = z_j \cdot \left[\Theta(-m_j)\cdot 2\sigma\left(\frac{m_j}{\tau}\right) + \Theta(m_j)\right].
\end{equation}
All the gating parameters are initialized to zero, such that we guarantee that, at initialization, the performance of the pruned model exactly matches the vanilla one. Then, the proxy parameters are updated according to two loss terms:
\begin{itemize}
    \item a task loss, wanted as low as possible to maintain performance on the target task;
    \item an empirical mutual information loss, on the predictions of on the bias estimator head, to be minimized as well.
\end{itemize}
Given the above, we can formulate an objective function to be minimized:
\begin{equation}
    \label{eq:optim}
    J = \mathcal{L}\left(y, \hat{y}\right) + \gamma \mathcal{I}\left(b, \hat{b}\right),
\end{equation}
where $\mathcal{I}\left(b, \hat{b}\right)$ is the empirical mutual information calculated for the given minibatch and $\gamma$ is a regularization coefficient. The update for $m_i$ follows standard back-propagation rules.

\subsection{Overview of FFW}
\algdef{SE}[DOWHILE]{Do}{doWhile}{\algorithmicdo}[1]{\algorithmicwhile\ #1}%
\begin{algorithm}[h]
\caption{Finding fantastic weights.}
\label{alg:ffw}
\begin{algorithmic}[1]
\Procedure{FFW($\ee$, $\cc$, $\mathcal{D}_{\text{train}}$, $\mathcal{D}_{\text{val}}$, mode)}{}
\State $m_i\gets 1 \forall i$\Comment{Initialize gating weights}
\State $\tau\gets 1$
\If{mode = unstructured}
\State Attach parameter mask estimator \eqref{eq:hatw} to $\ee$
\Else
\State Attach neuron mask estimator \eqref{eq:hatz} to $\ee$
\EndIf
\State Initialize bias extraction head $\pp$ and attach to the output of $\ee$
\Do
\Do
\State Train $\pp$ on $\mathcal{D}_{\text{train}}$
\State Optimize gating parameters $m_i$ according to \eqref{eq:optim} on $\mathcal{D}_{\text{train}}$
\doWhile{Plateau in performance on $\mathcal{D}_{\text{val}}$} 
\State $\tau \gets 0.5\cdot \tau$\Comment{Scale the temperature}
\doWhile{performance of model with $\tau$ is the same as $\tau=0$} 
\EndProcedure
\end{algorithmic}
\end{algorithm}

FFW is synthesized in Alg.~\ref{alg:ffw}. In our approach, we hypothesize that, given a sufficiently trained and parametrized network (namely, not in an under-fitting regime), a set of spurious and target features are learned and blended together. To this end, we target to learn a mask on the encoder $\ee$ such that the information on the bias is filtered and not usable by the task classifier $\cc$. We therefore initialize the gating parameters $m_i$ to $1$ and replace the parameters in $\ee$ by the expression \eqref{eq:hatw}, where $w_i$ is non-trainable but the only learnable parameter is $m_i$. The temperature $\tau$ is initialized to $1$, and a bias extraction head $\pp$ is initialized and attached to the output of $\ee$. Then, $\pp$ and the gating parameters are trained on the training set $\mathcal{D}_{\text{train}}$ until a plateau on a validation set $\mathcal{D}_{\text{val}}$ is reached. At this point, $\tau$ is rescaled by a factor $2$ and the process is iterated until the performance of the model matches the one of the model obtained with $\tau=0$, which is our stop criterion.We will provide our empirical findings below.

\section{Experiments}
\label{sec:experiments}

\subsection{Experimental Setup}
\label{sed:setup}

For our experiments, we employ architectures typically used for benchmarking bias in the four datasets Biased MNIST, CelebA, Corrupted CIFAR10 and Multi-Color MNIST (described in Sec.~\ref{sed:datasets}). Specifically, for Biased MNIST, we employ the same fully convolutional network used for~\cite{rebias}, composed of four convolutional layers with $7\!\times\!7$ kernels, with batch normalization layers. The training procedure to get the vanilla model is the same as in~\cite{irene}. For the experiments on CelebA and Corrupted-CIFAR10, we used a pre-trained Resnet-18 on ImageNet, and we employed the same optimization strategy as in~\cite{LfF,biasCon}. For our FFW, we employ in all our experiments $\gamma=10$ and we use a fixed learning rate for $\pp$ of $0.1$, until plateauing on the validation set. Noteworthily, we will employ two splits of the training set: one, biased, used for the vanilla training and indicated as $\mathcal{D}^{b}$, and another, unbiased and used by FFW to extract unbiased sub-networks, indicated as $\mathcal{D}^{u}$. In all our tables, we report three values for our experiments: ``Task'' indicates the top-1 accuracy on the target task, ``Bias'' indicates the performance of $\pp$ and acts as an upper bound on the information related to the bias information (as top-1 accuracy) usable by $\cc$, and the sparsity of the extracted sub-network (sometimes displayed as ``Spars.''). Our code is accessible at \hyperlink{https://github.com/renahon/debiasing_surgeon_ffw}{this link}.

\subsection{Datasets}
\label{sed:datasets}
Below, we provide an overview of the datasets used for quantitative evaluation, chosen to represent various biases types (synthetic or real-world) and number.
As stated in \ref{sed:setup}, we chose datasets that provide bias labels and train our gating weights on balanced, unbiased datasets.\\
\textbf{Biased MNIST.} Introduced by Bahng \textit{et al.}~\cite{rebias}, comprises 60k samples where MNIST handwritten digits are colored with a correlation $\rho$ between color and digits. Each digit is assigned a specific color, and the samples receive a background color accordingly. We test four levels of color-digit correlation: $\rho$ = \{0.99, 0.995, 0.997, and 0.999\}. The bias effect, namely the background color, is assessed by testing on an unbiased dataset: $\rho$ = 0.1. As this dataset can be generated artificially by injecting a background color to black-and-white digit, for the following experiments, we built two 60k datasets for each experiment: the biased $\mathcal{D}^b$ with a correlation $\rho$ and an unbiased $\mathcal{D}^u$ that we split in three parts of proportions 60\%-20\%-20\% to obtain our $\mathcal{D}^u_{\text{train}}, \mathcal{D}^u_{\text{val}}, $ and $\mathcal{D}^u_{\text{test}}$.\\
\textbf{CelebA.} CelebA~\cite{CelebA}, a real-world dataset commonly used for debiasing evaluations, consists of 203k face images annotated with 40 attributes. Here, we focus on the classification of whether the individual has  "BlondHair" or not, where the primary bias stems from the strong tendency for perceived females to be blond. \\
\textbf{Corrupted CIFAR10.} Similarly to Biased MNIST, Corrupted CIFAR10 consists in the injection of a specific correlation between an attribute (here an image corruption, such as "motion blur", or "fog") and one of the ten classes of CIFAR10. The corruptions are taken from \cite{CIFAR10c} and the corruption protocol was taken from \cite{barbano2023unbiased}. We tested their levels of correlation: $\rho$ = \{0.95, 0.98, 0.99, 0.995\}.\\
\textbf{Multi-Color MNIST.} To complexify the challenge posed by Biased MNIST, Li \textit{et al.} proposed a bi-colored version in~\cite{Debian} to assess models' performance on multiple biases simultaneously. In this version, the left and right sides of the background have different colors, each correlated to the target with $\rho_L$ and $\rho_R$ respectively. We adopt their setup with $\rho_L$ = 0.99 and $\rho_R$ = 0.95.

\subsection{Preliminary analysis}
\begin{table}[h]
\centering
\caption{Results for FFW applied on Biased-MNIST ($\rho=0.99$) for different $\gamma$.}
\label{tab:ffw_results_biased_mnist_gamma}
\resizebox{\textwidth}{!}{
\begin{tabular}{lcccccccc}
\toprule
\multirow{2}{*}{\textbf{Metric}}& \multicolumn{6}{c}{\textbf{Regularization coefficient $\gamma$}}\\
  & 2 & 5 & 10 & 20 & 50 & 200\\
\midrule
Task Accuracy
& $98.17 \pm 0.27$ & $98.07\pm 0.24$ & $97.79\pm 0.30$  & $97.52\pm 0.10$  & $93.68\pm 3.77$ & $23.31\pm 19.57$  \\
Bias Accuracy
& $21.49\pm5.68$ & $15.94\pm1.99$ & $15.74\pm3.26$ & $17.10\pm1.93$ & $16.02\pm0.58$ & $10.61\pm1.25$ \\
Sparsity
& $39.72\pm0.03$ & $41.49\pm2.75$ & $46.50\pm0.01$ & $46.90\pm0.02$ & $51.81\pm0.02$ & $52.13\pm0.03$ \\
\bottomrule
\end{tabular}
}%
\end{table}

We propose here a preliminary analysis where we analyze the impact of $\gamma$ on both the target task and the bias. We conduct this analysis on Biased-MNIST in the unstructured pruning setup. The results are reported in Tab.~\ref{tab:ffw_results_biased_mnist_gamma} and they are averaged on three seeds. As expected, we observe that for extremely large values of $\gamma$ the performance on the target task drops significantly, as the sparsity increases. However, we also notice that for values until $\gamma=20$ the task accuracy remains high. The bias accuracy remains consistently below 20\% for $\gamma>2$, which drives our selection to an intermediate value of $\gamma=10$ for the other experiments.

\subsection{Main results}
\label{sec:mainresults}
In this section, we discuss our results on the datasets introduced in Sec.~\ref{sed:datasets}.

\begin{table}[h]
\centering
\caption{Results for Biased-MNIST with different correlation levels.}
\label{tab:ffw_results_biased_mnist_rho}
\resizebox{\textwidth}{!}{
\begin{tabular}{lcccccccccccc}
\toprule
\multirow{3}{*}{\textbf{Method}}          & \multicolumn{12}{c}{\textbf{Proportion of bias-aligned samples: $\rho$}} \\
                                 & \multicolumn{3}{c}{0.99} & \multicolumn{3}{c}{0.995} & \multicolumn{3}{c}{0.997} & \multicolumn{3}{c}{0.999} \\
                                 \cmidrule{2-13}
                                 & Task  & Bias  & Spars.  & Task  & Bias  & Spars.   & Task  & Bias  & Spars. & Task   & Bias   & Spars. \\
\midrule
Vanilla                          & 88.46 & 98.45 & -          & 74.23 & 99.75 & -          & 46.04 & 99.96 & -         & 10.02  & 100.0 & -       \\
\midrule
Rubi \cite{rubi}                 & 93.6  & -     & -          & 43.0  & -     & -          & 90.4  & -     & -         & 13.7   & -      & -       \\
EnD \cite{end}                   & 96.0  & -     & -          & 93.9  & -     & -          & 83.7  & -     & -         & 52.3   & -      & -       \\
BCon+BBal \cite{biasCon}         & \textbf{98.1 } & -     & -          & \textbf{97.7}  & -     & -          & \textbf{97.3 } & -     & -         & \textbf{94.0 }  & -      & -       \\
ReBias \cite{rebias}             & 88.4  & -     & -          & 75.4  & -     & -          & 65.8  & -     & -         & 26.5   & -      & -       \\
LearnedMixin \cite{LearnedMixin} & 88.3  & -     & -          & 78.2  & -     & -          & 50.2  & -     & -         & 12.1   & -      & -       \\
LfF \cite{LfF}                   & 95.1  & -     & -          & 90.3  & -     & -          & 63.7  & -     & -         & 15.3   & -      & -       \\
SoftCon \cite{biasCon}           & 95.2  & -     & -          & 93.1  & -     & -          & 88.6  & -     & -         & 65.0   & -      & -       \\
\midrule
FFW Structured                   & 97.63 & \textbf{16.89} & 45.78      & 96.68 & \textbf{19.03} & 47.07      & 92.46 & 22.59 & \textbf{38.40}     & 19.91  & 17.57  & \textbf{47.24}   \\                
FFW                              & 97.79 & 15.74 & \textbf{46.50 }     & 97.30 & 17.14 & 44.62      & 89.36 & \textbf{27.84} & 36.47     & 22.29  & \textbf{23.13 } & 43.11 \\
                               
\bottomrule
\end{tabular}
}%
\end{table}

Tab.~\ref{tab:ffw_results_biased_mnist_rho} reports the results obtained for Biased-MNIST at different proportions of bias-misaligned samples (indicated as $\rho$). For the lower values of $\rho$ we can clearly notice that both FFW structured and unstructured are among other state-of-the-art approaches, suggesting that proper pruning of vanilla-trained models, without further retraining, can yield well-generalizing performance. For $\rho=0.999$, however, we observe that FFW is falling behind other approaches. This can be explained by the fact that the scarcity of unbiased representants in the training set prevents the vanilla model from learning the bias-disentangled information. In Sec.~\ref{sec:fitting} we display the need to properly fit the training data. 

\begin{table}[h]
\centering
\caption{Results for FFW applied on CelebA with different target attributes.}
\label{tab:ffw_results_celeba_correlations}
\resizebox{.65\textwidth}{!}{
\begin{tabular}{lcccccc}
\toprule
\multirow{3}{*}{\textbf{Method}}& \multicolumn{6}{c}{\textbf{Target task attribute}} \\
 & \multicolumn{3}{c}{BlondHair} & \multicolumn{3}{c}{Glasses} \\
 \cmidrule{2-7}
 & Task & Bias & Sparsity & Task & Bias & Sparsity \\
\midrule
Vanilla
           & 80.42 & 72.22 & 0.00 & 98.42 & 50.00 & 0.00 \\
\midrule
         EnD \cite{end}  &86.9 &-&-&-&-&-\\
         LNL \cite{LNL}  &80.1&-&-&-&-&-\\
         DI \cite{DI}  &90.9&-&-&-&-&-\\
         BCon+BBal \cite{biasCon}  & 91.4 &-&-&-&-&-\\
         Group DRO \cite{groupdro}  & 85.4&-&-&-&-&-\\
         LfF \cite{LfF}  & 84.2 &-&-&-&-&-\\
         LWBC \cite{LWBC}  & 85.5 &-&-&-&-&-\\
 \midrule
  FFW Structured & 86.95 & 48.00 & 49.11 & 98.28 & 49.35 & 46.43 \\
 FFW & 87.08 & 50.00 & 49.11 & 98.58  & 48.66 & 46.58\\
\bottomrule
\end{tabular}
}%
\end{table}

Tab.~\ref{tab:ffw_results_celeba_correlations} reports the results on the CelebA dataset for both the attributes ``BlondHair'' and ``Glasses'' assuming the gender being the bias. Regarding the first attribute, we observe that FFW stands among some debiasing algorithms, yielding performance around 87\% for both the structured and the unstructured variants. Some approaches, like BCon+BBal~\cite{biasCon} perform careful contrastive learning on the bias features weighted by the loss, and can achieve better performance; however, notably the vanilla model, properly pruned, ranks among the firsts. Interestingly we have reported the performance for the attribute ``Glasses'', uncommon for the debiasing community, to show that, despite dataset imbalances, some attributes are naturally ignored by vanilla training (the bias extraction on the vanilla model is already at random guess).

\begin{table}[h]
    \centering
    \caption{Results of FFW method for debiasing Corrupted-CIFAR10.}
    \label{tab:ffw_results}
\resizebox{\textwidth}{!}{
    \begin{tabular}{lcccccccccccc}
    \toprule
    \multirow{3}{*}{\textbf{Method}}& \multicolumn{12}{c}{\textbf{Proportion of bias-aligned samples: $\rho$}} \\
    & \multicolumn{3}{c}{0.995} & \multicolumn{3}{c}{0.99} & \multicolumn{3}{c}{0.98} & \multicolumn{3}{c}{0.95} \\
    \cmidrule{2-13}
    & Task & Bias & Spars. &  Task & Bias & Spars. & Task & Bias & Spars. &  Task & Bias & Spars. \\  
    \midrule

                     Vanilla                                  & 20.50  & 89.67   & 0.00       & 23.90  & 85.27 & 0.00       & 29.92  & 82.18 & 0.00      & 43.03   & 79.22  & 0.00 \\
            \midrule
            EnD~{\scriptsize\cite{end}}        & 19.38  & -       & -      & 23.12  & -     & -      & 34.07  & -     & -      & 36.57   & -      & -  \\
            HEX~{\scriptsize\cite{wang2018learning}}          & 13.87  & -       & -      & 14.81  & -     & -      & 15.20 & -     & -      & 16.04   & -      & -\\
            ReBias~{\scriptsize\cite{rebias}}        & 22.27  & -       & -      & 25.72  & -     & -      & 31.66  & -     & -      & 43.43   & -      & - \\
            LfF~{\scriptsize\cite{nam2020learning}}           & 28.57  & -       & -      & 33.07  & -     & -      & 39.91  & -     & -      & 50.27   & -      & -  \\
            DFA~{\scriptsize\cite{lee2021learning}}           & 29.95  & -       & -      & 36.49  & -     & -      & 41.78  & -     & -      & 51.13   & -      & - \\
            uLA~{\scriptsize\cite{grouprob}}           & 34.39  & -       & -      & \textbf{62.49}  & -     & -      & \textbf{63.88}  & -     & -      & \textbf{74.49}  & -      & - \\
            \midrule
FFW                                               & 51.57  & \textbf{10.03}  &\textbf{ 6.93}     & 58.25  &\textbf{9.77 }  & \textbf{2.73}    & 58.80   & 10.85 & \textbf{3.19 }   & 61.48   & 10.47  & \textbf{5.32} \\
FFW Structured                                    & \textbf{52.02}  & 10.65  & 5.26     & 57.92  &10.38  & 2.33    & \textbf{59.20 }&\textbf{ 10.43} & 3.14    & 60.97   & \textbf{10.40} & 5.31  \\

    \bottomrule
    \end{tabular}
}%
    \end{table}
    
Tab.~\ref{tab:ffw_results} reports the results on Corrupted-CIFAR10. In this case, the vanilla model pruned through FFW surpasses state-of-the-art approaches, by a high margin in some cases: for $\rho=0.05$, FFW beats other techniques by approximately 17\%. We hypothesize that the corruptions present in this dataset are easy to remove by pruning; still, it is surprising that the vanilla model, properly pruned and without extra fine-tuning, can reach such performance.
\begin{table}[h]
\centering
\caption{Results for FFW applied on Multi-Color MNIST.
The accuracies are computed on four subsets of the dataset. The ``Unbiased'' one is the average of the four. $A$ and $C$ stand for Aligned and Conflicting (regarding the samples).}
\label{multimnist}
\resizebox{\textwidth}{!}{
\begin{tabular}{lccccccccc}
\toprule
\multirow{2}{*}{\bf Method} 
& \multirow{2}{*}{\bf Sparsity [\%] ($\uparrow$)} 
& \multicolumn{2}{c}{\bf Bias acc. [\%] ($\downarrow$)} 
& \multicolumn{5}{c}{\bf Task acc. [\%] ($\uparrow$)}  \\       
  &      &Left& Right & \small{$A_{\text{L}}$ / $A_{\text{R}}$}  & \small{$A_{\text{L}}$ / $C_{\text{R}}$}  & \small{$C_{\text{L}}$ / $A_{\text{R}}$}          & \small{$C_{\text{L}}$  / $C_{\text{R}}$}      & \small{Unbiased} \\\midrule 
Vanilla            & 0.00       &96.38& 64.7               & 100.0        & 97.7        & 30.37        & 7.68         & 58.9                      \\
\midrule
LfF \cite{LfF}          & -     &-&-              & 99.6        & 4.7         & 98.6        & 5.1         & 52.0                      \\
EIIL \cite{EIIL}      & -         &-&-            & 100.0        & 97.2        & 70.8        & 10.9        & 69.7                      \\
PGI  \cite{PGI}       & -        &-&-                      & 98.6                    & 82.6        & 26.6        & 9.5         & 54.3                      \\
DebiAN  \cite{Debian}      & -     &-&-                & 100.0        & 95.6        & 76.5        & 16.0        & 72.0                      \\
VCBA \cite{VCBA}          & -    &-&-                        & 100.0 & 90.9  & 77.5  & 24.1  & $\boldsymbol{73.1} $               \\
\midrule
FFW  & 16.95 &  20.27 & 16.55 & 34.57& 35.17 & 39.86 & $\boldsymbol{35.85}$ & 36.37\\
\bottomrule
\end{tabular}%
}%
\end{table}

Finally, Tab.~\ref{multimnist} displays our results on Multi-Color MNIST. If FFW yields worst results on average than the vanilla network, we can notice that it led to the best results on the worst sub-group: the one with bias-conflicting backgrounds on both sides: we get an increase of 11\%. We can also notice that the results are comparable on every subset, meaning that the network doesn't base its choices on the background color anymore.

\subsection{Importance of fitting the training set}
\label{sec:fitting}
\begin{table}[htbp]
\centering
\caption{Results for Biased-MNIST with $\rho=0.99$ with different extraction times. Note that one full training epoch consists of 600 batches (therefore the three first columns represent respectively 10, 100, and 300 iterations). }
\label{tab:ffw_results_biased_mnist_extraction}
\resizebox{\textwidth}{!}{
\begin{tabular}{llcccccc}
\toprule
\multirow{2}{*}{\textbf{Method}} & \multirow{2}{*}{\textbf{Metric}}  & \multicolumn{6}{c}{\textbf{Epochs}}\\
& & $1/60$ & $1/6$ & $1/2$                  & 1 & 10 & 40\\
\midrule
\multirow{2}{*}{Vanilla} 
& Task      & $9.65\pm 0.24$&  $10.06\pm0.16$ &  $9.97\pm0.01$ &  $9.96\pm0.01$ &$16.68\pm3.25$ & $89.34\pm0.45$\\
& Bias   & $100.0\pm 0.0$&  $100.0\pm 0.0$ &  $100.0\pm 0.0$ &  $100.0\pm 0.0$ & $100.0\pm 0.0$ & $98.69\pm0.36$ \\
\midrule
\multirow{3}{*}{FFW} 
& Task     
& $9.71 \pm 0.68$ & $11.16\pm0.33$ & $10.30\pm0.91$ & $12.42\pm2.86$ & $46.39\pm18.70$  & $97.96\pm0.30$\\
& Bias   
& $9.98\pm 0.08$ & $10.03\pm 0.00$ & $10.02\pm 0.02$ & $13.48\pm 6.03$ & $23.81\pm 10.65$ & $15.81\pm 1.64$ \\
& Sparsity   
& $50.34\pm 0.13$ & $51.41\pm 0.52$ & $50.86\pm 0.36$ & $51.19\pm 0.28$ & $48.26\pm 2.48$ & $42.01\pm 5.40$ \\
\bottomrule
\end{tabular}
}
\end{table}

As an ablation study, we propose to extract unbiased sub-networks from a vanilla-trained model from models trained for a different amount of epochs. For this study, we select Biased-MNIST with $\rho=0.99$. We report the results in Tab.~\ref{tab:ffw_results_biased_mnist_extraction}, averaging the results on three seeds. For this setup, we evidently observe that, even training less than an epoch (as little as 1/60 of an epoch) the vanilla model perfectly fits the bias, and the disentangled features are completely not learned, until the completion of the first epoch. Progressively though, non-biased features are learned by the model, and the accuracy of the target task increases. This suggests that a simple approach as FFW can work on vanilla-trained models if they are sufficiently trained (and fitting) on the target task.   

\iffalse
\begin{table}[htbp]
\centering
\caption{Results for Biased-MNIST ($\rho = 0.99$ with different pruning strategies.}
\label{tab:ffw_results_multicolor_mnist}
\begin{tabular}{cccc}
\toprule
Strategy &Sparsity &\multicolumn{2}{c}{Accuracy} \\ 
        &           & Task & Bias \\
\midrule
Vanilla Model & 0
&  $88.46\pm0.63$ &  $98.45\pm0.20$  \\
\midrule
\multirow{5}{*}{Magnitude Pruning} 
& 0.5
&  $88.06\pm0.8$ &  $99.04\pm0.17$  \\

& {0.75}
& $78.16\pm4.59$ &  $99.58\pm0.14$  \\

& {0.9}
& $31.14\pm9.96$ &  $98.30\pm0.23$  \\

& {0.95}
& $10.50\pm1.16$ &  $77.45\pm11.29$  \\

& {0.975}
& $9.52\pm0.53$ &  $40.98\pm7.18$  \\
\midrule
{FFW Structured} &{0.46}
  & $97.63\pm1.02$     & $16.89\pm3.15$\\
{FFW} & {0.47} 
& $97.79\pm0.30$ & $15.64\pm3.26$\\
\bottomrule
\end{tabular}
\end{table}
\fi

\section{Conclusion}
\label{sec:conclusion}

In this work, we have analyzed the problem of finding, from vanilla-trained models on known biased setups, debiased sub-networks that can both solve the target task and not rely on the information related to the bias itself. Such an analysis is very important for the debiasing community for both the increasing concerns related to trustworthiness in AI systems and efficiency in treating models already trained. Indeed, the typical approach followed by the debiasing community is to fine-tune the whole model to remove such a source of bias, showcasing enhanced performance in unbiased environments.

Our first contribution to this work lies in the fact that removing the bias does not necessarily improve the final task's performance. Indeed, in the case such a feature identified as ``bias'' is inherently necessary to solve a target task, its removal leads to a performance drop. We have derived a theoretical framework that also suggests such a behavior.

Secondly, we have proposed FFW, a technique that, without requiring fine-tuning the vanilla model, is able to surgically remove parameters from the model, unveiling the existence of an unbiased sub-network. We have proposed both an unstructured and a structured variant of such an approach, which also provides guarantees on the biased information employable for solving the target task. On four common benchmarks, we have observed that such sub-networks exist, leading to performance comparable with other state-of-the-art approaches. One major finding is that they are even structured, potentially leading also to computational gains. This finding bridges the sparsity and debiasing communities, opening the road to the design of more energy-efficient debiasing approaches.

\section*{Acknowledgements}
 Part of this work was funded by the French National Research Agency (ANR-22-PEFT-0007) as part of France 2030 and the NF-FITNESS project and by Hi!PARIS Center on Data Analytics and Artificial Intelligence. Besides, this work
was also funded by the European Union’s Horizon Europe research and innovation program under grant agreement No.
101120237 (ELIAS).

\bibliographystyle{splncs04}
\bibliography{main}

\appendix
\addcontentsline{toc}{section}{Appendices}
\section*{Appendices}
% \documentclass[runningheads]{llncs}

% \usepackage{eccv}

% \usepackage{eccvabbrv}

% \usepackage{multirow}
% \usepackage{multicol}
% \usepackage{graphicx}
% \usepackage{booktabs}
% \usepackage{algorithm}
% \usepackage{algpseudocode}
% \usepackage{enumitem}
% \usepackage{wrapfig}

% \usepackage[accsupp]{axessibility}  % 
% \def\xx{\mathbf{x}}
% \def\ee{\mathcal{E}}
% \def\cc{\mathcal{C}}
% \def\pp{\mathcal{P}}
% \def\w{\mathbf{w}}
% \def\i{\mathcal{I}}
% \def\l{\mathcal{L}}
% \def\epsun{\varepsilon_{\text{unb}}}
% \def\epsbia{\varepsilon_{\text{bia}}}
% \def\kk{K_{\text{bia}}}
% \newcommand{\parallelsum}{\mathbin{\|}}

% \usepackage{hyperref}

% \usepackage{orcidlink}

% \begin{document}

% % ---------------------------------------------------------------
% \title{Debiasing surgeon: fantastic weights and how to find them -- Appendix} 

% \author{Rémi Nahon\orcidlink{0009-0009-0757-5716} \and
% Ivan Luiz De Moura Matos\orcidlink{0009-0009-5032-0444}\and
% Van-Tam Nguyen \and Enzo Tartaglione\orcidlink{0000−0003−4274−8298}}

% \authorrunning{R.~Nahon et al.}

% \institute{LTCI, Télécom Paris, Institut Polytechnique de Paris, France 
% \email{\{name.surname\}@telecom-paris.fr}}

% \maketitle
% \appendix
% \setcounter{equation}{10}
% \setcounter{table}{6}
% \setcounter{figure}{3}

\section{Visualizations}
\subsection{Is FFW helping the network focus on the right features ?}
\begin{figure}
    \centering
    \includegraphics[width=0.5\linewidth]{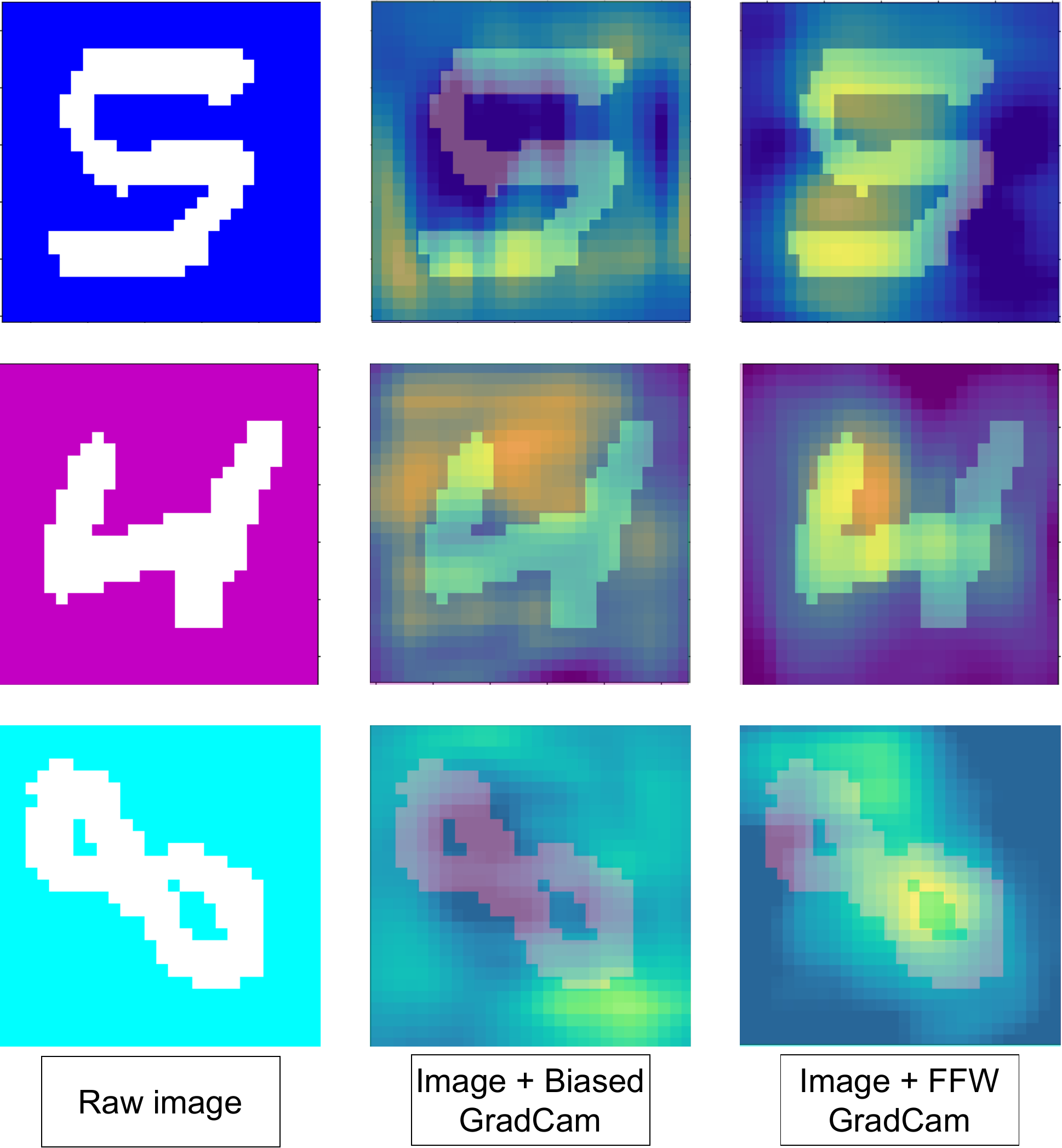}
    \caption{Grad-Cam visualization of the effects of FFW on Biased-MNIST with $\rho = 0.997$}
    \label{fig:gradcam}
\end{figure}
For Fig.\ref{fig:gradcam}, we applied the Grad-Cam visualization method from \cite{gradcam} to FFW on Biased-MNIST. The raw sample (on the left) leads to high activation around the digit for the biased model (in the middle) but after pruning with FFW, the higher activations are placed on the digit.
\subsection{Pruning distribution across the networks.}
\begin{figure}[htbp]
  \centering
  \begin{tabular}{cc}
    \includegraphics[width=0.5\textwidth]{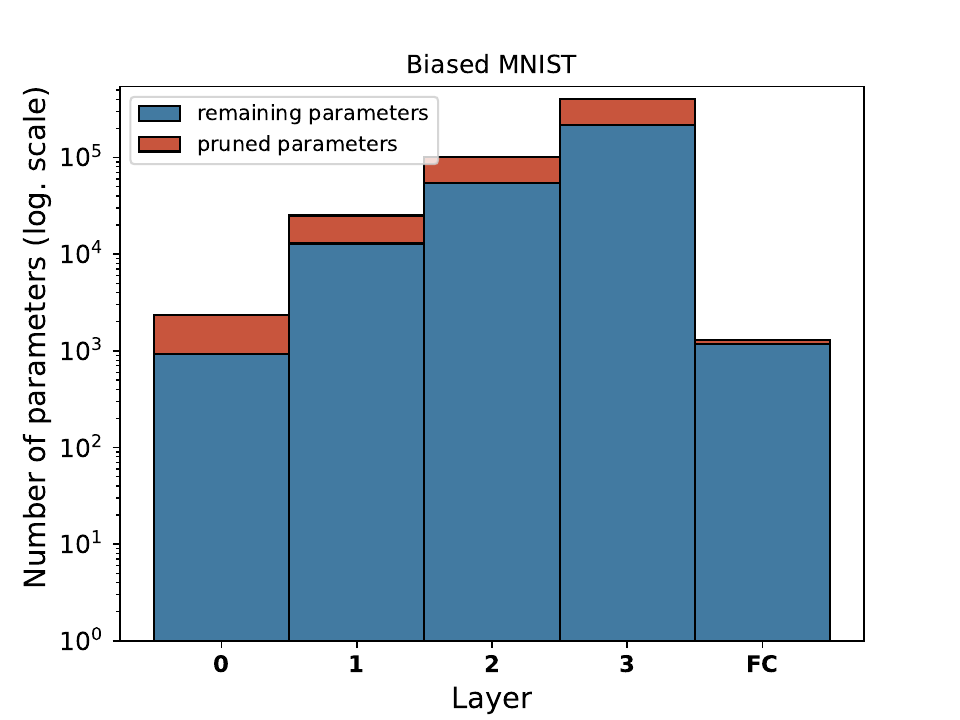} & 
        \includegraphics[width=0.5\textwidth]{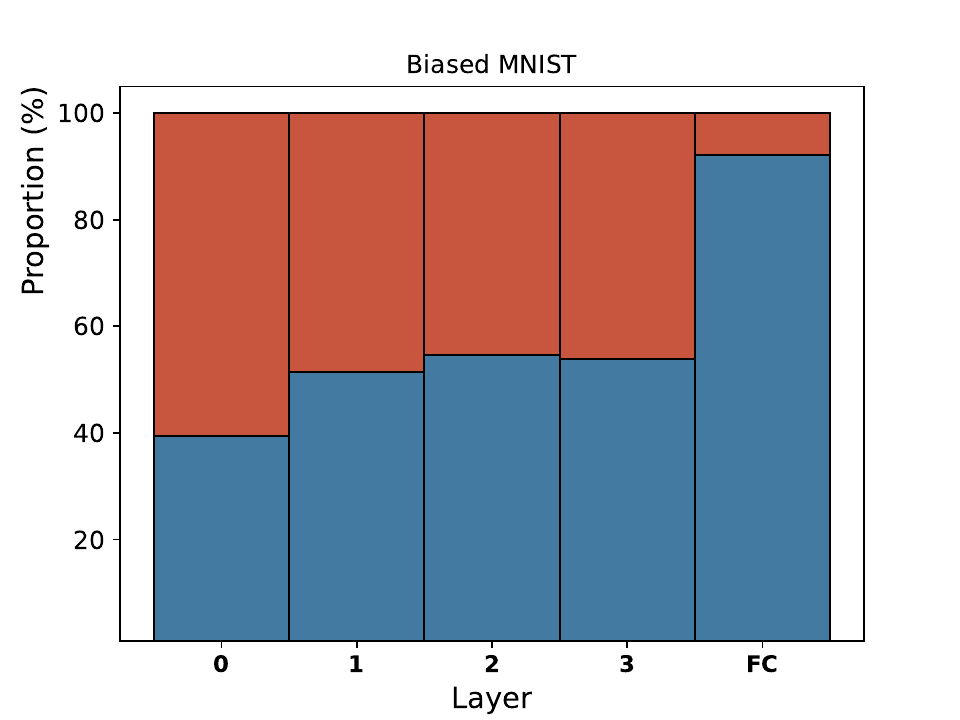}
    \\
 \includegraphics[width=0.5\textwidth]{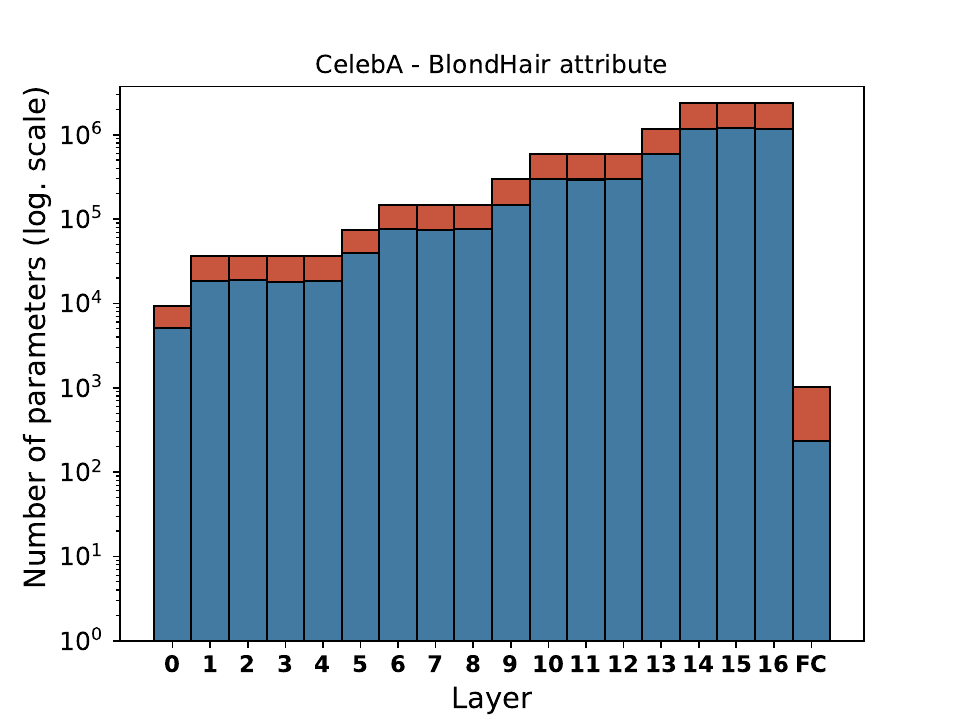} & 
        \includegraphics[width=0.5\textwidth]{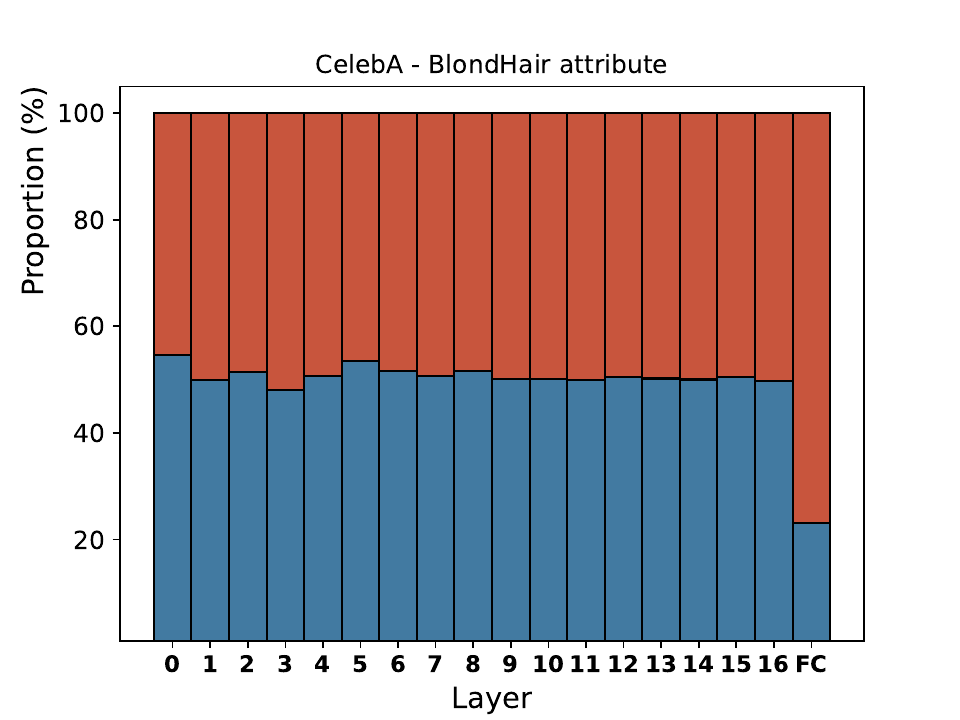}
        \\
        \includegraphics[width=0.5\textwidth]{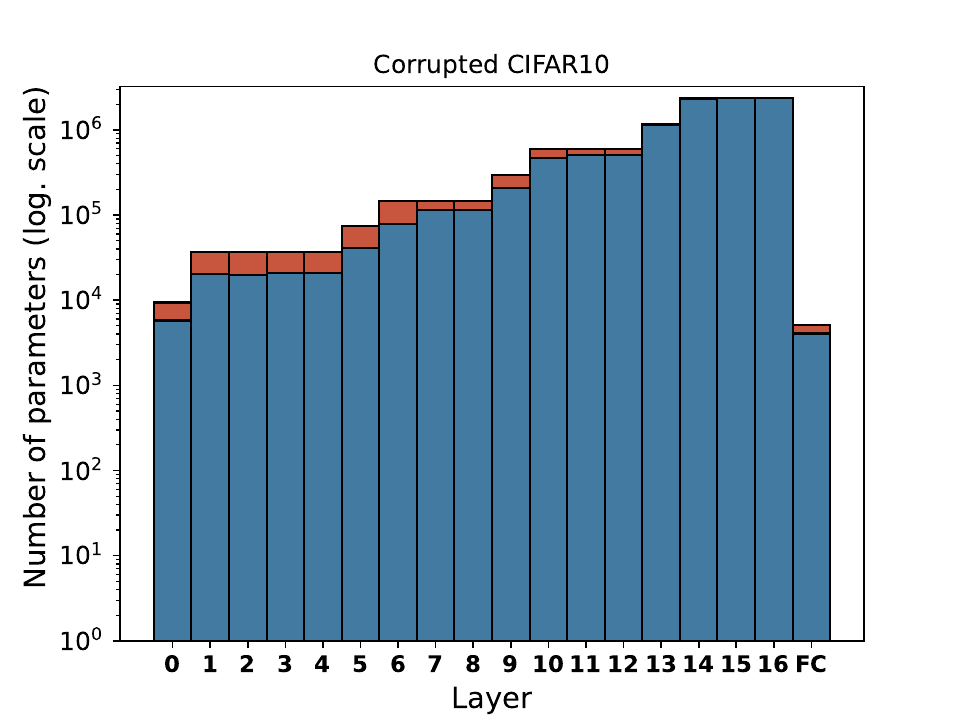} & 
        \includegraphics[width=0.5\textwidth]{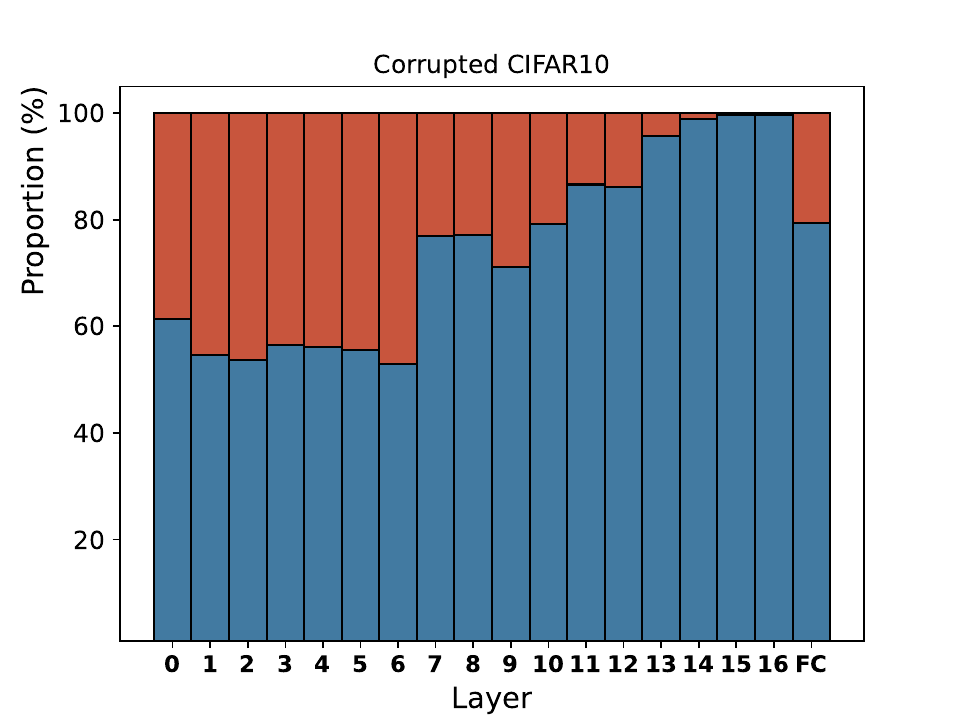}
  \end{tabular}
  \caption{Absolute number (top row) and proportions (bottom row) of pruned parameters after applying FFW to Biased MNIST, CelebA, and Corrupted CIFAR10.}
  \label{fig:pruning}
\end{figure}
Fig.~\ref{fig:pruning} shows the proportions of pruning in multiple networks for multiple datasets and two networks. It should be noted that while the network's pruning is spread over all layers for CelebA, it is focused on the first few layers for Corrupted Cifar10, potentially indicating the higher simplicity of the bias (a simple filter applied to the images) for that dataset.
\subsection{Variations on the Mutual Information Minimization}
\begin{figure}
    \centering
    \includegraphics[width=0.8\linewidth]{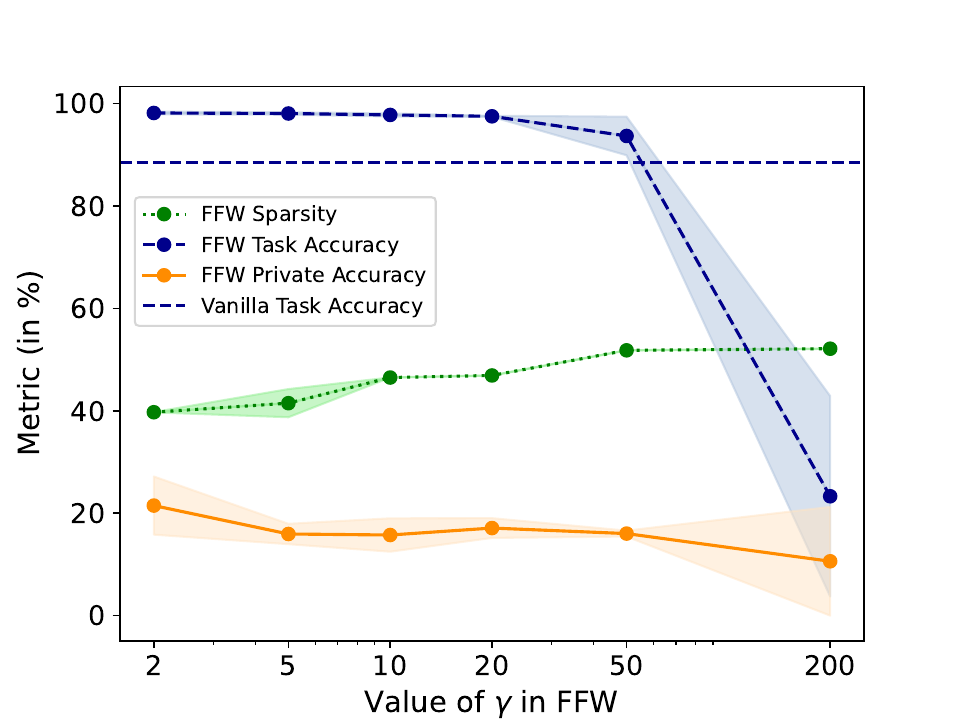}
    \caption{Results for FFW applied on Biased-MNIST ($\rho= 0.99$) with different $\gamma$ on the validation set}
    \label{fig:gamma}
\end{figure}
In Fig.~\ref{fig:gamma} we can visualize the results of Tab. 1, showing that the Task Accuracy can be maintained high while minimizing the Private Accuracy for $\gamma$ ranging from 5 to 50 on that specific dataset.
\subsection{Comparison of Pruning Strategies}
\begin{table}[htbp]
\centering
\caption{Results for Biased-MNIST ($\rho = 0.99$) with different pruning strategies.}
\label{tab:pruning}
\begin{tabular}{cccc}
\toprule
Strategy &Sparsity &\multicolumn{2}{c}{Accuracy} \\ 
        &           & Task & Bias \\
\midrule
Vanilla Model & 0
&  $88.46\pm0.63$ &  $98.45\pm0.20$  \\
\midrule
\multirow{5}{*}{Magnitude Pruning} 
& 0.5
&  $88.06\pm0.8$ &  $99.04\pm0.17$  \\

& {0.75}
& $78.16\pm4.59$ &  $99.58\pm0.14$  \\

& {0.9}
& $31.14\pm9.96$ &  $98.30\pm0.23$  \\

& {0.95}
& $10.50\pm1.16$ &  $77.45\pm11.29$  \\

& {0.975}
& $9.52\pm0.53$ &  $40.98\pm7.18$  \\
\midrule

{FFW Structured} &{0.46}
  & $97.63\pm1.02$     & $16.89\pm3.15$\\

{FFW} & {0.47} 
& $97.79\pm0.30$ & $15.64\pm3.26$\\
\bottomrule
\end{tabular}
\end{table}
\begin{figure}
    \centering
    \includegraphics[width=0.8\linewidth]{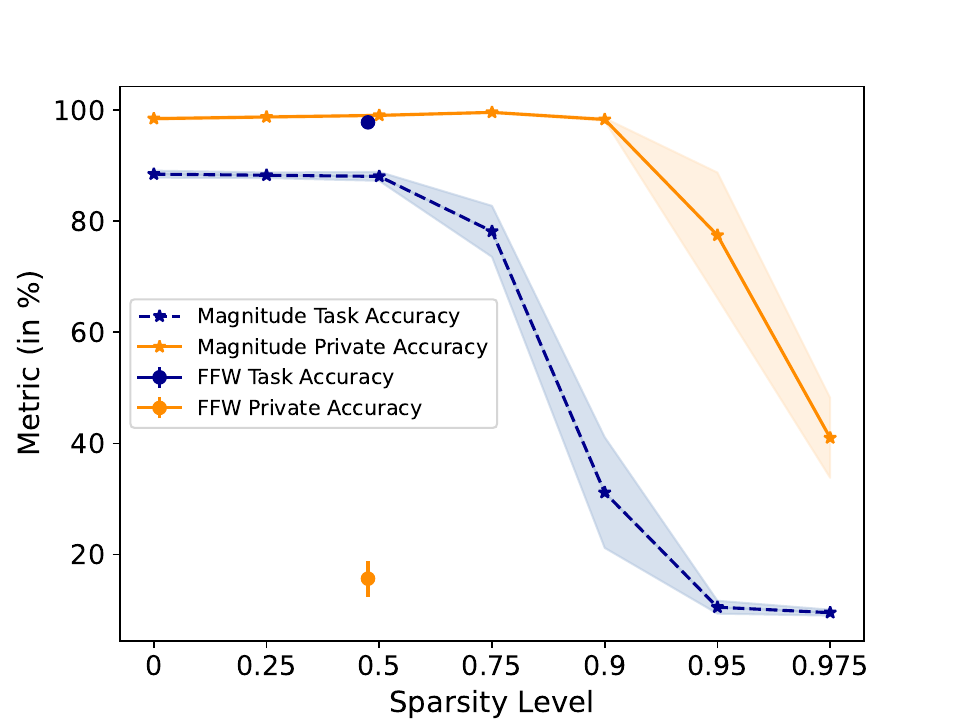}
    \caption{Results of Magnitude Pruning on Biased MNIST ($\rho=0.99$) at different levels of sparsity compared to FFW.}
    \label{fig:pruning_strat}
\end{figure}
In Fig.~\ref{fig:pruning_strat}, we can compare FFW to a vanilla magnitude pruning approach. While the magnitude pruning approach destroys both the Task and the Bias information while pruning, which is clear in the fact the both curves drop as sparsity increase, FFW does the opposite. At its 46$\%$ of sparsity, the pruning leads to a Private Accuracy close to random guess while increasing the Task Accuracy.
\subsection{Effect of the extraction point}
\begin{figure}
    \centering
    \includegraphics[width=0.8\linewidth]{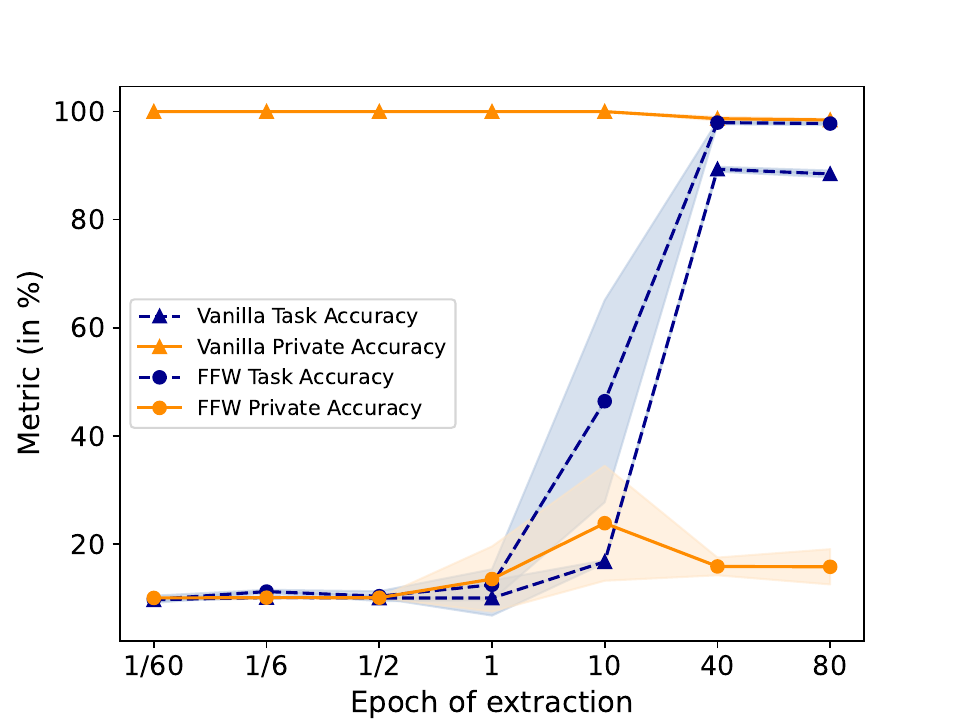}
    \caption{Results of FFW on Biased MNIST ($\rho=0.99$) applied after different biased training durations.}
    \label{fig:extraction}
\end{figure}
Fig.~\ref{fig:extraction} helps us visualize the results of Tab.5. Indeed it shows that until the completion of at least a few epochs, the model fits only the bias and therefore, it is impossible to extract a debiased subnetwork by our method. However, as the network learns unbiased features, our method keeps minimizing the propagation of biased information but starts progressively extracting subnetworks that are efficient on the target task. 
\subsection{Training times}
We provide the average training times of our method (after training the vanilla model) in Tab.~\ref{tab:training_time}.
\begin{table}[t]
\centering
\caption{Training time for our main experiments performed on a NVIDIA A100-PCIE-40GB GPU.}
\label{tab:training_time}
\resizebox{\columnwidth}{!}{
\begin{tabular}{cccc}
\toprule
Dataset           & Model used            & Training time & Unbiased training set \\
\midrule
Biased-MNIST \cite{rebias}    & SimpleConvNet \cite{rebias}  & 3h20  & 36000                          \\
CelebA \cite{CelebA}          & ResNet18                        & 1h15 & 437                         \\
Corrupted CIFAR10 \cite{CIFAR10c} & ResNet18                        & 0h52 & 6000                     \\
Multi-Color MNIST \cite{Debian} & 3-layers MLP & 0h22 & 6000\\
\bottomrule
\end{tabular}
}
\end{table}
This duration varies with the sizes of the unbiased training set, of the input images as well as the model type.
\section{Details for Section 3}
In this section, we will provide all the derivations for Sec.~3 in the main paper. more specifically, Sec.~\ref{sec:deriv3} will propose the derivation for (3), Sec.~\ref{sec:joint} will discuss the joint probability that will be employed to derive (4) in Sec.~\ref{sec:4deriv} and (6) in Sec.~\ref{sec:6deriv}.
\subsection{Derivation for (3)}
\label{sec:deriv3}
Given the joint probability as in (2), we can easily express the mutual information between $\hat{B}$ and $\hat{Y}$ as
\begin{align}
    \mathcal{I}(\hat{B},\hat{Y}) =& \sum_{i,j} p(\hat{b}_j,\hat{y}_i) \log_2\frac{p(\hat{b}_j,\hat{y}_i)}{p(\hat{b}_j)p(\hat{y}_i)}\nonumber\\
    =&\frac{N_B}{N_C}\rho \log_2\left[\frac{\rho N_C N_B}{N_C}\right] + \frac{N_B(N_B - 1)}{N_C(N_B-1)}(1-\rho) \log_2\left[\frac{(1-\rho)N_C N_B}{N_B-1}\right]\nonumber \\
    =& \frac{N_B}{N_C} \left\{ \rho \log_2(N_B) + \rho \log_2(\rho) + (1-\rho) \log_2(N_B) + (1-\rho)\log_2 \left( \frac{1-\rho}{N_B -1}\right)\right\}\nonumber\\
    =&\frac{N_B}{N_C} \left\{ \log_2(N_B) + \rho \log_2(\rho) + (1-\rho)\log_2 \left( \frac{1-\rho}{N_B -1}\right)\right\}\label{eq:3final}
\end{align}

\subsection{Joint probability between $\hat{B}$, $\hat{Y}$, $Y$}
\label{sec:joint}
A clear dependency between $\rho$ and (3), as already showcased in Sec.~\ref{sec:deriv3}, exists. This measure is applied to ground-truth labels, investigating the common information between them (and for instance, the information that is possible to disentangle). Nonetheless, in the more general case, the trained model (whose output is modelizable as the random variable $Y$) is not a perfect learner, having $H(Y|\hat{Y}) \neq 0$.
The model, in this case, does not correctly classify the target for two reasons.
\begin{enumerate}
	\item It gets confused by the bias features, and it tends to learn to classify samples based on them. We model this tendency of learning biased features with $\phi$, which we call \emph{biasedness}. The higher the biasedness is, the more the model relies on features that we desire to suppress, inducing bias in the model and for instance error in the model.
	\item Some extra error $\varepsilon$, non-directly related to the bias features, which can be caused, for example, by stochastic unbiased effects, to underfit, or to other high-order dependencies between data. This contribution is already visible in (1).
\end{enumerate}
We can write the discrete joint probability for $\hat{B}, \hat{Y}, Y$, composed of the following terms.
\begin{itemize}
    \item When target, bias, and prediction are aligned, the bias is aligned with the target class and correctly classified. Considering that we are not perfect learners, we introduce the error term $\varepsilon$.
    \item When the target and bias are misaligned and the prediction is correct, it means that the model has learned the correct feature and the bias is being contrasted. This effect is due to the dual effect of both the model's biasedness $\phi$ and the inherent ground-truth dependence between the use of the biased feature and the target label $\kk$.
    \item When target and bias are not aligned, but the prediction is incorrect and bias and output are aligned, it means that the model has learned the bias, introducing the error we target to minimize in this work.
    \item In all the other cases, the error of the model is due to higher-order dependencies, not directly related to the biasedness $\phi$.
\end{itemize}
Under the assumption $~{N_B=N_C=N}$, we can write the joint distribution
\begin{align}
p(\hat{B},\hat{Y},Y) =& 
\frac{1}{N} \cdot \left[
\delta_{\hat{b}\hat{y}y} \rho (1 - \varepsilon) 
+ \delta_{\hat{y}y}\bar{\delta}_{\hat{b}y} \bar{\delta}_{\hat{b}\hat{y}} \frac{(1-\phi)(1-\rho)}{N-1}(1-\kk)+ \nonumber \right .\nonumber \\%[1em]
	&
 + \bar{\delta}_{\hat{y}y} \delta_{\hat{b}y} \bar{\delta}_{\hat{b}\hat{y}} \frac{\phi(1-\rho)}{N-1}\kk 
 + \bar{\delta}_{\hat{y}y}\bar{\delta}_{\hat{b}y} \delta_{\hat{b}\hat{y}} \frac{\varepsilon \rho^2}{N-2+\rho}+ \nonumber \\%[1em]
	&\left . 
 + \bar{\delta}_{\hat{y}y}\bar{\delta}_{\hat{b}y}\bar{\delta}_{\hat{b}\hat{y}} \frac{\varepsilon \rho (1-\rho)}{(N-1)(N-2+\rho)} \right].\label{eq:jointbyy}
\end{align}

\subsection{Derivation for (4)}
\label{sec:4deriv}
Let the joint probability as in \eqref{eq:jointbyy}. We can marginalize on $\hat{Y}$ by summing all the $N_C$ (in our simplified case, $N$) biases per given target class and prediction of the model:
\begin{equation}
    p(\hat{B},Y) = \frac{1}{N}\left\{\delta_{\hat{b}y}\left[\rho(1-\varepsilon) + \phi(1-\rho)\right] +\bar{\delta}_{\hat{b}y} \left[ \frac{(1-\phi)(1-\rho)}{N-1} + \frac{\rho\varepsilon}{N-2+\rho} \right]\right\}.
\end{equation}

\noindent From this, following the definition of mutual information, we can write
\begin{align*}
    \mathcal{I}(\hat{B},Y) =& \sum_{i,j} p(\hat{b}_j,y_i) \log_2\frac{p(\hat{b}_j,y_i)}{p(\hat{b}_j)p(y_i)}\nonumber\\
    =&\frac{1}{N} \left\{ N \cdot\left[\rho (1 - \varepsilon) + \phi(1-\rho) \right] \cdot \right. \log_2 (N \cdot(\rho (1 - \varepsilon) + \phi(1-\rho))) +\nonumber \\
    &+ (N^2 - N) \cdot \left [\frac{(1-\phi)(1-\rho)}{N-1} + \frac{\rho\varepsilon}{N-2+\rho}\right ]\cdot\nonumber\\
    &\left .\cdot \log_2 \left[ N \cdot \left(\frac{(1-\phi)(1-\rho)}{N-1} + \frac{\rho\varepsilon}{N-2+\rho}\right)\right ] \right\}\nonumber
\end{align*}
\begin{align*}
    =&\left[\rho (1 - \varepsilon) + \phi(1-\rho) \right] \cdot  \left[\log_2 N + \log_2(\rho (1 - \varepsilon) + \phi(1-\rho)\right]+ \nonumber\\
    &+ \left [(1-\phi)(1-\rho) + \frac{(N - 1)\rho\varepsilon}{N-2+\rho}\right ]\cdot\nonumber\\
    &\cdot \left[ \log_2 N + \log_2 \left(\frac{(1-\phi)(1-\rho)}{N-1} + \frac{\rho\varepsilon}{N-2+\rho}\right)\right ]\nonumber
\end{align*}
\begin{align}
    =&\log_2\left[\rho (1 - \varepsilon) + \phi(1-\rho)\right]^{\rho (1 - \varepsilon) + \phi(1-\rho)}\nonumber\\
    &+ \log_2 \left(\frac{(1-\phi)(1-\rho)}{N-1} + \frac{\rho\varepsilon}{N-2+\rho}\right)^{(1-\phi)(1-\rho) + \frac{(N - 1)\rho\varepsilon}{N-2+\rho}}\nonumber\\
    &+\log_2 N \left[\rho (1 - \varepsilon) + \phi(1-\rho) +\right.\nonumber\\
    &+\left .(1-\phi)(1-\rho) + \frac{(N - 1)\rho\varepsilon}{N-2+\rho}\right ]\nonumber\\
    =&\log_2\left[\rho (1 - \varepsilon) + \phi(1-\rho)\right]^{\rho (1 - \varepsilon) + \phi(1-\rho)} \nonumber\\
    &+ \log_2 \left(\frac{(1-\phi)(1-\rho)}{N-1} + \frac{\rho\varepsilon}{N-2+\rho}\right)^{(1-\phi)(1-\rho) + \frac{(N - 1)\rho\varepsilon}{N-2+\rho}}\nonumber\\
    &+\log_2 N \left[1-\rho\varepsilon + \frac{(N - 1)\rho\varepsilon}{N-2+\rho}\right ].\label{eq:IBZA}
\end{align}
From here, we can easily obtain the normalized mutual information by scaling down the results of a factor $\log_2(N)$. Under the assumption that $\varepsilon = 0$, we obtain
\begin{align}
    \mathcal{I}(\hat{B},Y) =& \frac{1}{\log_2(N)} \left\{(\rho + \phi(1-\rho))\log_{2}\left[\rho + \phi(1-\rho)\right]+\right.\nonumber\\
    &\left . +[(1-\phi)(1-\rho)]\log_{2} \left [\frac{(1-\phi)(1-\rho)}{N-1} \right ]+ 1 \right\}\nonumber\\
    =&\frac{\rho + \phi(1-\rho)}{\log_2(N)}\log_2\left[\rho + \phi(1-\rho)\right]+ \frac{(1-\phi)(1-\rho)}{\log_2(N)} \log_2 \left [\frac{(1-\phi)(1-\rho)}{N-1} \right ]+ 1.\label{eq:final4}
\end{align}
We apologize for the typos in (4), which will be corrected as in \eqref{eq:final4} in the final version of the paper.

\subsection{Derivation for (6)}
\label{sec:6deriv}
Similarly to the approach taken in Sec.~\ref{sec:4deriv}, from the joint probability as in \eqref{eq:jointbyy}, we marginalize, but this time on $\hat{B}$, by summing all the $N_B$ (in our simplified case, $N$) biases per given target class and prediction of the model. Under the assumption that $\varepsilon=\varepsilon_{\text{bia}}$, we have:
\begin{align}
    p(\hat{Y},Y) =& \frac{1}{N}\left\{\delta_{\hat{y}y}\left[\rho(1-\varepsilon_{\text{bia}}) + (1-\phi)(1-\rho)(1-\kk) \right] +\right.\nonumber\\
    &\left.+\bar{\delta}_{\hat{y}y} \left[ \frac{\phi(1-\rho)}{N-1}\kk + \frac{\varepsilon_{\text{bia}}\rho^2}{N-2+\rho} + \frac{N-2}{N-2+\rho}\frac{\varepsilon_{\text{bia}} \rho (1-\rho)}{N-1}\right]\right\}
\end{align}

Also in this case, following the definition of mutual information, we can write
\begin{align*}
    \mathcal{I}(\hat{Y},Y) =& \sum_{i,j} p(\hat{y}_j,y_i) \log_2\frac{p(\hat{y}_j,y_i)}{p(\hat{y}_j)p(y_i)}\nonumber\\
    =&\frac{1}{N} \left\{ N \cdot\left[\rho (1 - \varepsilon) + \phi(1-\rho) \right] \cdot  \log_2 (N \cdot(\rho (1 - \varepsilon) + \phi(1-\rho))) +\right.\nonumber \\
    &+ (N^2 - N) \cdot \left [\frac{(1-\phi)(1-\rho)}{N-1} + \frac{\rho\varepsilon}{N-2+\rho}\right ]\cdot  \nonumber\\
    &\log_2 \left[ N \cdot \left.\left(\frac{(1-\phi)(1-\rho)}{N-1} + \frac{\rho\varepsilon}{N-2+\rho}\right)\right ] \right\}\nonumber\\
    =&\left[\rho (1 - \varepsilon) + \phi(1-\rho) \right] \cdot \left[\log_2 N + \log_2(\rho (1 - \varepsilon) + \phi(1-\rho)\right]+ \nonumber\\
    &+ \left [(1-\phi)(1-\rho) + \frac{(N - 1)\rho\varepsilon}{N-2+\rho}\right ]\cdot\nonumber\\
    &\cdot \left[ \log_2 N + \log_2 \left(\frac{(1-\phi)(1-\rho)}{N-1} + \frac{\rho\varepsilon}{N-2+\rho}\right)\right ].
\end{align*}
Here, defining 
\begin{equation}
    f(x,y,z) = xy \log_2(xz),
\end{equation}
we can write
\begin{align*}
    \mathcal{I}&(\hat{Y}, Y) = f\left\{\frac{1}{N}\left[\rho(1- \epsbia) + (1-\phi)(1-\rho)(1-\kk)\right], N,N^2\right\} +\nonumber\\
    &f\left\{\frac{1}{N}\left[\frac{\phi(1-\rho)}{N-1}\kk + \left(\frac{\rho^2(N-2) + \rho(1-\rho)(N-2)}{(N-2+\rho)}\right)\epsbia\right], N(N-1), N^2\right\},\label{eq:MIdependence}
\end{align*}
finding back (6).
% \bibliographystyle{splncs04}
% \bibliography{main}
% \end{document}

\end{document}